\documentclass[]{spie}  

\usepackage{subfigure}

\usepackage{multicol}
\usepackage{multirow}
 
\usepackage{amsmath,amsfonts,amssymb}
\usepackage{graphicx}
\usepackage[colorlinks=true, allcolors=blue]{hyperref}

\title{A Comparative Study of High-Recall Real-Time Semantic Segmentation Based on Swift Factorized Network}
\author{Kaite Xiang, Kaiwei Wang$^{*}$ and Kailun Yang
}

\affil{State Key Laboratory of Modern Optical Instrumentation, Zhejiang University, No. 38 Zheda Road, Hangzhou 310027, China}
\authorinfo{* Corresponding author. Email: wangkaiwei@zju.edu.cn.}

\begin{document} 
\maketitle

\begin{abstract}
Semantic Segmentation (SS) is the task to assign a semantic label to each pixel of the observed images, which is of crucial significance for autonomous vehicles, navigation assistance systems for the visually impaired, and augmented reality devices. However, there is still a long way for SS to be put into practice as there are two essential challenges that need to be addressed: efficiency and evaluation criterions for practical application. For specific application scenarios, different criterions need to be adopted. Recall rate is an important criterion for many tasks like autonomous vehicles. For autonomous vehicles, we need to focus on the detection of the traffic objects like cars, buses, and pedestrians, which should be detected with high recall rates. In other words, it is preferable to detect it wrongly than miss it, because the other traffic objects will be dangerous if the algorithm miss them and segment them as safe roadways. In this paper, our main goal is to explore possible methods to attain high recall rate. Firstly, we propose a real-time SS network named Swift Factorized Network (SFN). The proposed network is adapted from SwiftNet, whose structure is a typical U-shape structure with lateral connections. Inspired by ERFNet and Global convolution Networks (GCNet), we propose two different blocks to enlarge valid receptive field. They do not take up too much calculation resources, but significantly enhance the performance compared with the baseline network. Secondly, we explore three ways to achieve higher recall rate, \textit{i.e} loss function, classifier and decision rules. We perform a comprehensive set of experiments on state-of-the-art datasets including CamVid and Cityscapes. We demonstrate that our SS convolutional neural networks reach excellent performance. Furthermore, we make a detailed analysis and comparison of the three proposed methods on the promotion of recall rate.
\end{abstract}


\section{INTRODUCTION}
\label{sec:intro}  

Semantic Segmentation (SS) is the task to assign a semantic label to each pixel of the observed images, which is of crucial significance for autonomous vehicles, navigation assistance systems for the visually impaired, and augmented reality devices~\cite{yang2018unifyingterrainawareness}\cite{xiang2019importance}. It has gained great development from traditional method into methods based on deep convolutional neural networks (CNNs) since the milestone created by Fully Convolutional Networks (FCN)~\cite{long2015fully}.

However, there is still a long way for SS to be put into practice as there are two essential challenges that need to be addressed. Firstly, the inference efficiency is paramount for real-time applications with limited computational resources, which should be assured. Secondly, different evaluation criterions have been used in the literature, without consideration of the target application scenario. To address these issues, we aim to lay a good balance between inference speed and performance measured using certain criterion suitable for certain applications instead of merely depending on mean Intersection over Union (mIoU).

In this paper, we propose a real-time SS network named Swift Factorized Network (SFN) based on ResNet-18~\cite{he2016deep}, a light-weight convolutional neural network specially designed for classification tasks. The proposed network is adapted from SwiftNet~\cite{orsic2019defense}, which can reach state-of the art performance at a fast inference speed. The structure of the network is a typical U-shape structure with lateral connections. Inspired by ERFNet~\cite{romera2018erfnet}\cite{romera2019bridging} and Global Convolutional Network~\cite{peng2017large}, we have realized the importance of the receptive field. In order to extract better features and enlarge valid receptive field, we propose two different blocks. The first method is to add a series of factorized convolution blocks with different dilation rate, and the other one is to insert global convolutional blocks with diverse kernel sizes to refine features. They won't take up too much calculation resources, but significantly enhance the performance compared with the baseline network.
   \begin{figure} [t]
   \begin{center}
   \begin{tabular}{c} 
   \includegraphics[height=6cm]{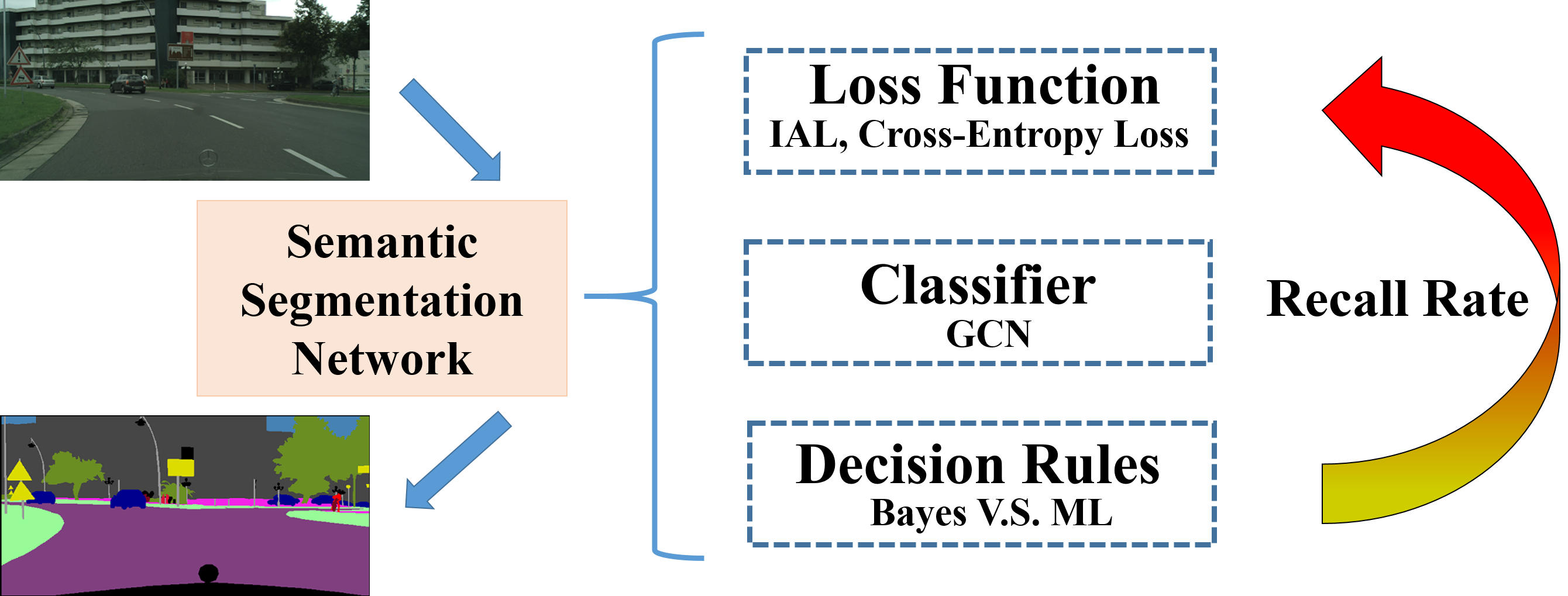}
   \end{tabular}
   \end{center}
   \caption[example] 
   { \label{fig:1}Basic procedure of the paper: a semantic segmentation network with three key methods towards high recall. }
   \end{figure}

On the other hand, in order to cope with specific application scenarios, we need to adopt different criterions. Recall rate is an important criterion for many tasks. Taking autonomous driving as an example, we need to focus on the detection of the traffic objects like cars, buses, and pedestrians, which should be detected with high recall rates. In other words, it is preferable to  detect it wrongly than miss it, because the other traffic objects will be dangerous if the algorithm miss them and segment them as safe roadways. Moreover, for autonomous driving, different objects should be coped with different rank of importance, the traffic light, cars, riders are much more important than road, sky, and etc. If the algorithm does not differentiate the different priorities for certain classes, the module will reach a normal performance for all classes. In order to promote the certain classes' recall rates, we explore three different methods as shown in Fig.~\ref{fig:1}:

(1)\textbf{ Loss Function:} Loss function is of great significance for training a model. In order to reach high recall rate, we utilize importance-aware loss~\cite{chen2018importance}.

(2)\textbf{ Classifier:} Graph Convolution Networks (GCN) are very useful and popular to cope with the graph data. It can also be utilized as a classifier, such that the result is based on the pre-defined graph.

(3)\textbf{ Decision Rules:} The normal decision rule in classification or semantic segmentation tasks is Bayes rule. In our experiment, we also adopt Maximum Likelihood (ML) rule which is promising to boost the recall rate.

In order to test and verify our method and network, we perform a comprehensive set of experiments on state-of-the-art datasets including CamVid~\cite{brostow2009semantic} and Cityscapes~\cite{cordts2016cityscapes}. Our contributions are summarized as follows: (1) A semantic segmentation network \textit{i.e.} SFN is proposed, which is a real-time network and can reach excellent performance. (2) We make a detailed analysis and comparison of the three proposed methods on the promotion of recall rate. (3) We provide three possible methods to attain higher recall rates for semantic scene understanding. (4) Our implementations and codes are available at: \url{https://github.com/Katexiang/swiftnet/tree/master/Swift_Factorized_Network(SFN)}
\section{RELATED WORK}

\subsection{Semantic Segmentation Neural Networks}
\label{sec:title}
The milestone created by FCN results in the prosper for method based on Deep Learning to SS task. The fundamental framework of SS Neural Networks is an Encoder-Decoder structure. Typical representative networks like UNet~\cite{ronneberger2015u}, PSPNet~\cite{zhao2017pyramid}, DeepLab V3~\cite{chen2017rethinking} and ACNet~\cite{hu2019acnet} can reach brilliant performance but can not strike a balance between efficiency and accuracy. Therefore, they can not be applied into real-time application. In order to put SS into practice, many real-time SS networks have been proposed in recent years. ENet~\cite{paszke2016enet} is the pioneer in the era of real-time SS networks, which is adapted from ResNet structure~\cite{he2016deep} but discards the last stage of the model to raise efficiency. Following that, many real-time SS Network appear. ERFNet~\cite{romera2018erfnet}\cite{romera2019bridging} and ERF-PSPNet~\cite{yang2018unifying}\cite{yang2019can} utilize residual factorized module to reduce parameters and keep fine performance. ICNet\cite{zhao2018icnet} and BiSeNet~\cite{yu2018bisenet} are multiply-path structures using immense calculations on small feature maps for excavating context information and a few calculations on big feature maps to keep spatial information so that they can keep high efficiency and favourable performance. SwiftNet~\cite{orsic2019defense} and many other real-time SS networks are based on the Encoder-Decoder structure with light-weight base networks like ShuffleNet\cite{zhang2018shufflenet}, MobileNet\cite{howard2017mobilenets} and etc., whose performance vary considerably depending on their decoder structures and lateral connections.

\subsection{Somewhat-Aware Method for Semantic Segmentation}
As is detailed in Sec.~\ref{sec:intro}, every pixel can not be addressed without bias. Inspired by the different attributes for different pixels, many somewhat-aware methods are proposed to cope with certain somewhat-bias tasks. Focal loss~\cite{lin2017focal} is one of the most well-known methods aiming to cope with the detection of difficult objects for object detection task, which has been transferred into SS task and can promote SS networks' precision~\cite{yang2018predicting}. Another difficulty-aware method is a structure style named deep layer cascade~\cite{li2017not} to cope with the situation when different pixels possess different difficulties. In view of the lack of valid label for SS, many researchers consider to utilize synthetic data to train SS Networks. However, the domain gap between synthetic data and real data prevents the models generalizing well to real image~\cite{romera2019bridging}. Therefore, Hierarchical Region Selection~\cite{sun2019not} is proposed to select similarity region between synthetic and real image, \textit{i.e.} an interest-aware method. IAL~\cite{chen2018importance} is a loss function which is put forward to cope with the fact that different objects possesses different ranks of importance, \textit{i.e.} an importance-aware method. IAL is one of the possible methods to achieve high-recall semantic segmentation.

\subsection{Graph Convolution Networks}
Graph Convolution Networks (GCN) are a kind of structure to generalize neural networks to graph data. For computer vision tasks, many method based on GCN have reached state-of-the-art performance in different areas. Their usuage can be classified into two types, \textit{i.e.} classifier and feature modifier. For the former,  Chen et al.~\cite{chen2019multi} utilize GCN to cope with multi-label image recognition and reach state-of-the-art performance. For the latter, Qi et al.~\cite{qi20173d} adopt GCN to act as an feature modifier to adjust the feature representation for RGBD SS. In previous works, we have also developed segmentation frameworks suitable for RGBD~\cite{hu2019acnet}\cite{yang2019robustifying} and panoramic images~\cite{yang2019can}. In this paper, GCN acts as a classifier in SS task, in order to advance recall rate depending on our pre-defined graph.

\subsection{Decision Rules}
There are generally two kinds of decision rules in machine learning task, \textit{i.e.}  Bayes rule and Maximum Likelihood (ML) rule. Standard decision rule for SS task is Bayes rule, while some research shows that Maximum Likelihood rule help advance the recall rate for SS task. Chan et al.~\cite{chan2019application} utilize ML rule to handle class imbalance in SS. ML is a promising key to reach high recall rate for SS.

\section{METHODOLOGY}
\subsection{Swift Factorized Network}
\label{sec:sfn} 
   \begin{figure} [ht]
   \begin{center}
   \includegraphics[height=5.2cm]{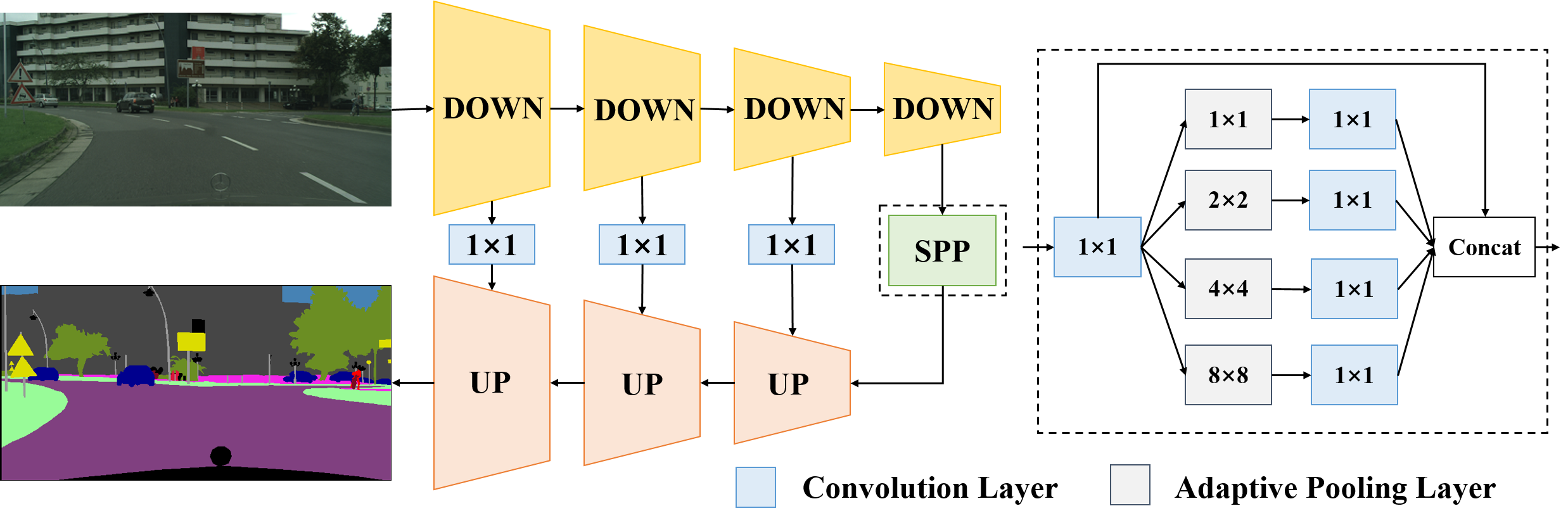}
   \end{center}
   \caption[example] 
   { \label{fig:2}Basic structure of SFN: the box in yellow is the Encoder, the box in orange is the Upsampling Decoder, the box in green is Spatial Pyramid Pooling and the dotted box is the detailed structure of SPP.}
   \end{figure} 
   
\begin{figure}[t] 
  \centering 
  \subfigure[Basic version]{ 
    \includegraphics[height=5cm]{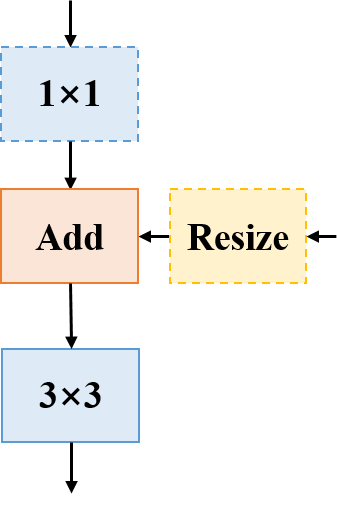} 
  } 
  \subfigure[ERF version]{ 
    \includegraphics[height=5cm]{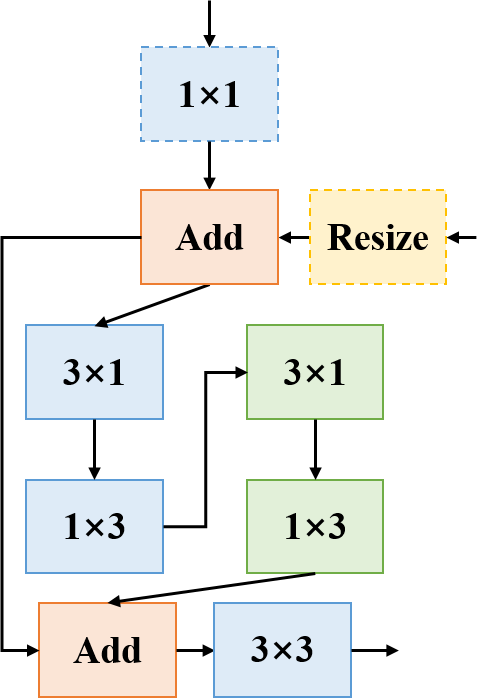} 
  } 
    \subfigure[Late merge of GCNet]{ 
    \includegraphics[height=5cm]{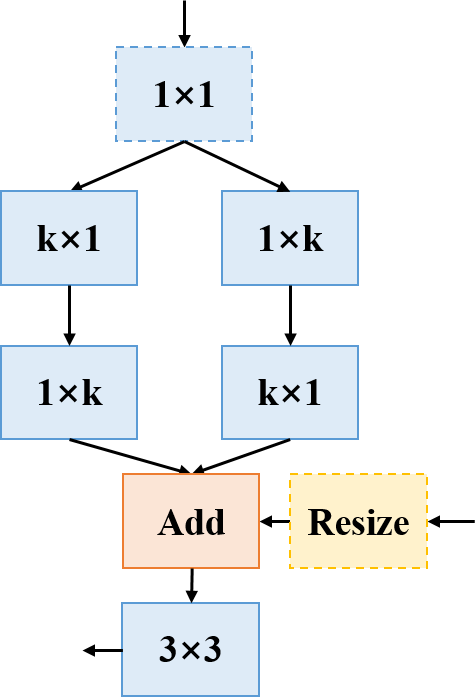} 
  } 
    \subfigure[Early merge of GCNet]{ 
    \includegraphics[height=5cm]{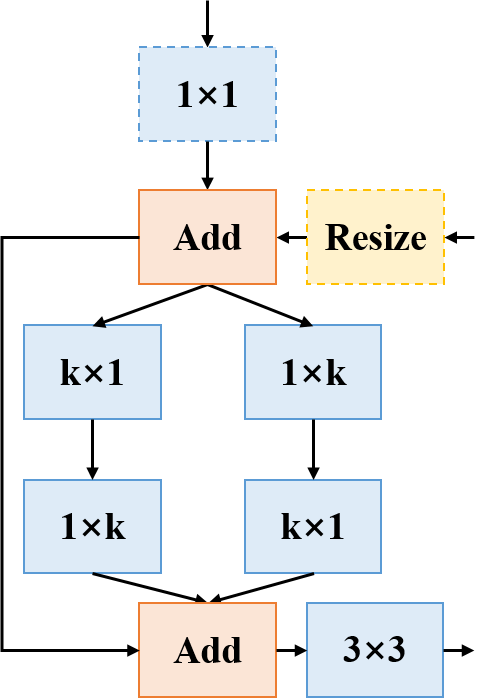} 
  } 
  \caption{\label{fig:3}The proposed Upsampling Decoder Blocks: (a) is the basic version of SwiftNet, (b) is the variation of ERFNet's ``Non-bottleneck-1D'' block where the boxes in green are the dilated convolution layers whose dilation rates will be studied in the experiment section. Additionally, (c) and (d) are the structure of the GCNet version, where `k' stands for the kernel size of the block. The bigger the kernel size is, the larger the receptive field is.} 
\end{figure}   
   
The proposed SS Network is based on the Encoder-Decoder structure with U-shape. The main architecture is shown in Fig.~\ref{fig:2}. The basic building blocks are Encoder, Spatial Pyramid Pooling (SPP) and Decoder which consists of a series of upsample blocks.

\textbf{Encoder:} For the downsample block, \textit{i.e.} the Encoder of the network, we choose ResNet-18 as the backbone of the SS network. The encoder working as the feature extractor can obtain semantic feature from the input image for SS, because it is brought up to cope with the recognition task. As a matter of fact, MobileNet and ShuffleNet are the alternatives as well. All of these models are compatible with real-time operation due to their small computation complexity. On the other hand, they are appropriate choices for fine tuning because there are many ImageNet~\cite{russakovsky2015imagenet} pre-trained parameters being publicly available.

\textbf{SPP:} SPP module is a typical pyramid structure which can extract feature for multi-scale representation, thus increasing the receptive field~\cite{he2015spatial}. As shown in the dotted box of Fig.~\ref{fig:2}, the spp module is composed of four pairs of convolution layers and adaptive pooling layer with varying scales or levels which can produce and recombine the feature maps at different scales and enlarge the valid receptive field. The structure is inherited from PSPNet, but transferred into a simplified version in order to advance efficiency.

\textbf{Upsampling Decoder Blocks (UDBs):} Traditional upsampling methods for SS include bilinear interpolation and transpose convolution. The upsampling decoder is to upsample feature maps from the Encoder to the input resolution. The original UDB is a lateral connection structure shown in Fig.~\ref{fig:3}(a), which is composed with a 1 $\times$ 1 convolution for dimension matching between feature maps from corresponding layers and those from previous layers, a bilinear interpolation layer for keeping the same resolution and a 3 $\times$ 3 convolution layer for blending and recombining features. This method is so simple that the valid receptive field is not very large only if the receptive field merely depends on SPP. Inspired by ERFNet, which utilizes a series of ``Non-bottleneck-1D'' block with various of dilation rates and factorized convolution, we redesign the base UDB into the structure as shown in Fig.~\ref{fig:3}(b). Factorized convolution is utilized to reduce parameters and dilation rate is employed to enlarge receptive field. According to GCNet, we have realized the significance of large kernel which can enlarge the receptive field directly. We also modify the base UDB into two additional versions as shown in Fig.~\ref{fig:3}(c) and Fig.~\ref{fig:3}(d). Their main difference is their merging stage between feature maps from corresponding layers and those from previous layers. We will detail the comparison among these four versions of UDBs, and choose the early merge of GCNet as the baseline block in the end named Swift Factorized Network (SFN).

\subsection{Loss function}
Loss function for our experiment includes cross-entropy loss and importance-aware loss. In fact, the latter one is the variation of cross-entropy loss. 

\textbf{The cross-entropy} is defined by
\begin{equation}
\label{eq:cross}
\textbf{I} = -\sum_{}\textbf{\textit{q}}\cdot log(\textbf{\textit{p}}) \, ,
\end{equation}
where \textbf{\textit{q}} and \textbf{\textit{p}} are the the one-hot encoding label and output after softmax-layer, respectively. The loss function is  widely used in the region of classification and SS task. In order to cope with the situation that different objects take up different pixel counts in the images. Frequency weighting \textbf{\textit{w}} is introduced into the loss function, and the modified cross-entropy loss becomes: 
\begin{equation}
\label{eq:cross2}
\textbf{\textit{w}}= {1\over \ln(\textbf{a}+\textbf{\textit{f}})} \, ,\textbf{I} = -\sum_{}\textbf{\textit{w}}\cdot\textbf{\textit{q}}\cdot log(\textbf{\textit{p}}) \, ,
\end{equation}
where \textbf{\textit{f}} is the frequency of the calculated class, and \textbf{a} is a hyper-parameter avoiding divided by zero. In this paper, we set it to 1.02. However, cross-entropy loss can not cope with the fact that different objects possess different ranks of importance.

\begin{figure}[h] 
  \centering 
  \subfigure[Importance rank]{ 
    \includegraphics[height=3cm]{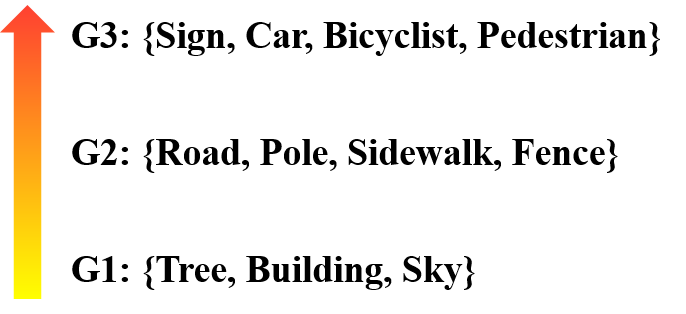} 
  } 
  \subfigure[\textbf{M$_{1}$}]{ 
    \includegraphics[height=3cm]{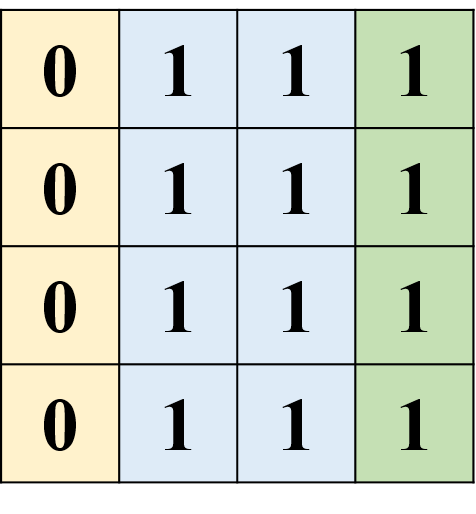} 
  } 
    \subfigure[\textbf{M$_{2}$}]{ 
    \includegraphics[height=3cm]{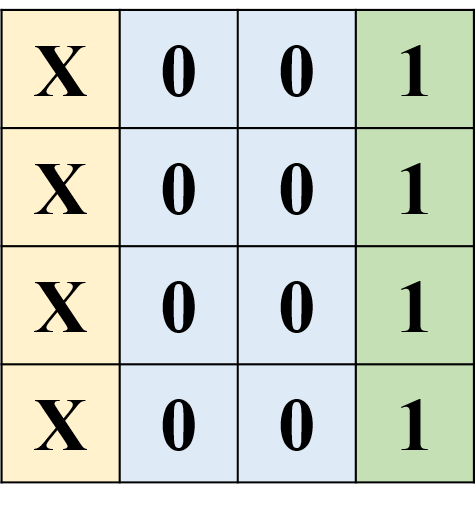} 
  } 

  \caption{\label{fig:4} (a) illustrates the importance ranking of classes for CamVid, where G1 is the most important group. (b) and (c) are the importance matrices where the green area is G3, blue area is G2, and yellow area is G1.} 
\end{figure}

\textbf{IAL} is a loss function, which is combined with cross-entropy loss and importance weight. Taking  CamVid  as  an  example,  the  dataset  has  11  classes,i.e., sky, building, pole, road, sidewalk, tree, sign, fence, car,pedestrian and bicyclist. In view of autonomous vehicles, there are three ranks of importance as shown in Fig.~\ref{fig:4}(a).
Fig.~\ref{fig:4}(b) and Fig.~\ref{fig:4}(c) are the importance matrices for calculating the importance dynamic weights. The importance dynamic weights are defined by: 
\begin{equation}
\label{eq:ial1}
\textbf{\textit{f$_{1}$}}={\sum{[(\textbf{M$_{1}$}+\lambda)^{0.5}\cdot (\textbf{\textit{p}}^{\prime}-\textbf{M$_{1}$})]^2}} \, ,
\textbf{\textit{f$_{2}$}}={\sum{[(\textbf{M$_{2}$}+\lambda)^{0.5}\cdot (\textbf{\textit{p}}^{\prime}-\textbf{M$_{2}$})\cdot(\textbf{M$_{2}$}\ne \textbf{X})]^2}} \, ,
\end{equation}
where $\textbf{\textit{p}}^{\prime}$ is the ground-truth channel value of the output and the value of ($\textbf{M$_{2}$}\ne \textbf{X}$) is 0 if the value of the matrix is \textbf{X} else the value is 1. $\lambda$ is a tuning parameter set to 0.5 in order to take the lower-importance category into consideration and avoid ignoring them when calculating the dynamic importance weight. The proposed \textbf{IAL} can be defined by
\begin{equation}
\label{eq:ial2}
\textbf{IAL} = \textbf{I$_{1}$}+(f_{1}+\alpha)\cdot\textbf{I$_{2}$}+(f_{2}+\alpha)\cdot(f_{3}+\alpha)\cdot\textbf{I$_{3}$}  \, ,
\end{equation}
where \textbf{I$_{1}$}, \textbf{I$_{2}$} and \textbf{I$_{3}$} are the sum of G1,G2 and G3's cross-entropy loss, respectively.

\subsection{Graph Convolution Network as Classifier}
Graph convolution networks are very popular in various research fields. In this paper, we make a brief introduction to GCN. Unlike normal convolution layer which operates on images or feature maps as a sliding window, GCN is an operation to extract relation among nodes in a graph. $\boldsymbol{H}^{l} \in \mathbb{R}^{n \times d}$ is a feature representation for a graph, where \textbf{\textit{l}} is the layer of the GCN, \textbf{\textit{n}} denotes the number of nodes and \textbf{\textit{d}} is the dimension of node features. The following layer of the GCN will be $\boldsymbol{H}^{l+1} \in \mathbb{R}^{n \times d^{\prime}}$. The calculation process can be represented as:
\begin{equation}
\label{eq:gcn}
\boldsymbol{H}^{l+1}=h\left(\widehat{\boldsymbol{A}} \boldsymbol{H}^{l} \boldsymbol{W}^{l}\right)  \, ,
\end{equation}
where $h(\cdot, \cdot)$ is an activation function (we select LeakyReLU in our experiments), $\widehat{\boldsymbol{A}}\in \mathbb{R}^{n \times n}$ is the normalized version of correlation matrix, and
$ \boldsymbol{W}^{l} \in \mathbb{R}^{n \times n}$
$\boldsymbol{H}^{l} \in \mathbb{R}^{d \times d^{\prime}}$ is the transformation matrix.

\begin{figure} [t]
   \begin{center}
   \includegraphics[height=5.2cm]{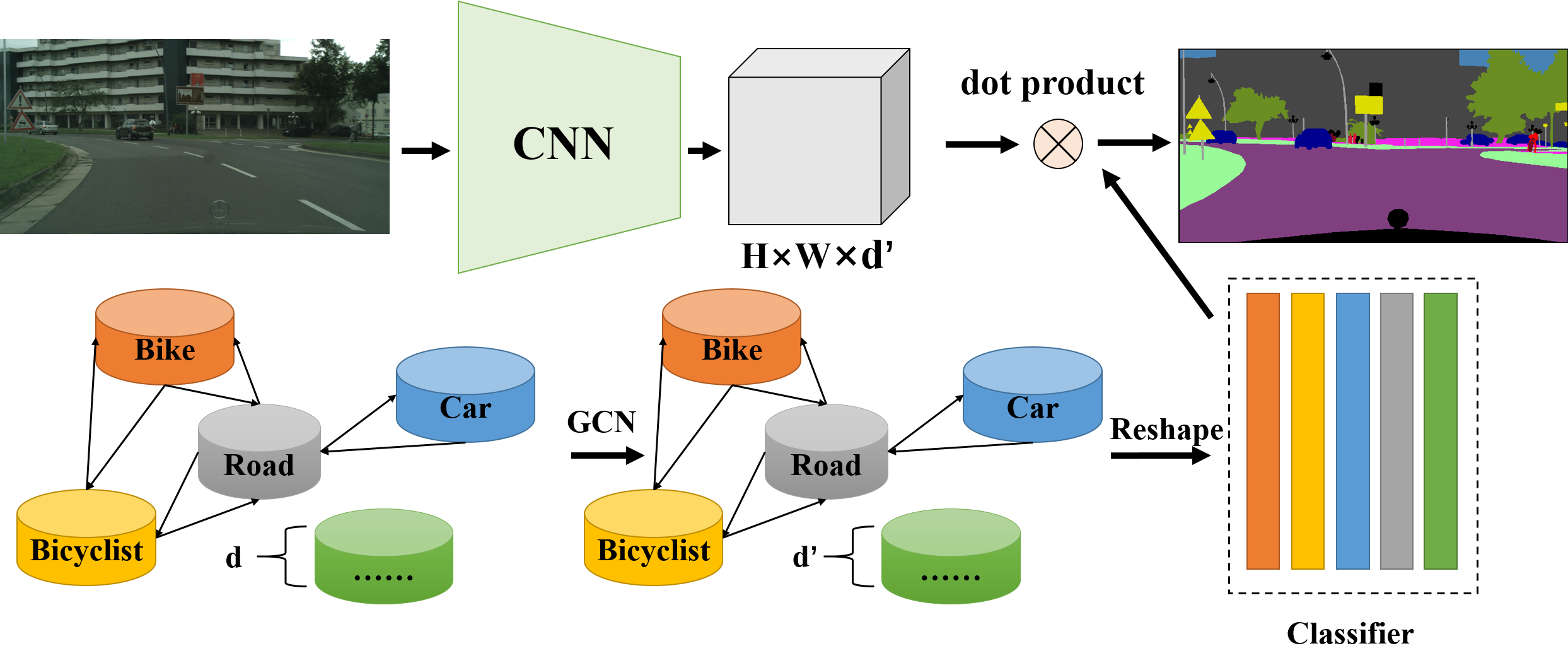}
   \end{center}
   \caption[example] 
   { \label{fig:5}Overall framework of SFN based on GCN classifier: the bottom of the figure is GCN and the dotted box is the classifier to select meaningful features.}
   \end{figure}

In our experiments, GCN works as an classifier as shown in Fig.~\ref{fig:5}. We define a directed graph, and utilize the graph to extract the relation among the classes. Then, the final layer of the GCN is reshaped into a matrix which works as the feature selector, \textit{i.e.} classifier.
 \begin{figure}[t] 
  \centering 
  \subfigure[pixel-wise priors for bicycle]{ 
    \includegraphics[height=4cm]{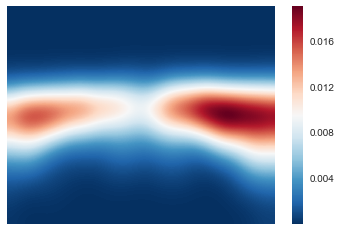} 
  } 
  \subfigure[pixel-wise priors for rider]{ 
    \includegraphics[height=4cm]{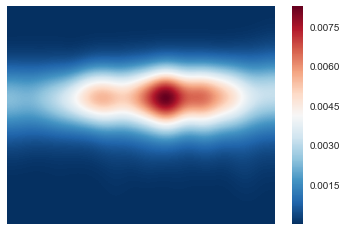} 
  } 

  \caption{\label{fig:6}The pixel-wise priors for bicycle and rider.} 
\end{figure}

\subsection{Decision Rule for Semantic Segmentation}
\label{sec.dr}
Standard decision rule for SS is Bayes rule. Normally, the output of the SS Network is followed by a softmax layer, which makes the sum of the output's every channel be 1. In other words, the output can be viewed as the probability of belonging to one of these categories. The decision result of the Bayes rule is that the maximum of the probability, \textit{i.e.,}
\begin{equation}
\label{eq:bayes}
d_{Bayes}(x)=\arg \max \,p(k|x),
\end{equation}
and Maximum Likelihood (ML) is defined by:
\begin{equation}
\label{eq:ml}
d_{M L}(x)={\arg \max } \,p(x | k){=} {\arg \max } \,p(k | x) p(x) / p(k)={\arg \max } \,p(k | x) / p(k),
\end{equation}
where $p(k)$ is the distribution of the ground truth labels, \textit{i.e.} priors, which is the key element to the implementation of ML decision. We estimate the priors by using the training set. More precisely, we need to calculate pixel-wise class distribution, in order to reduce data-specific noise from the priors, a Gaussian filter will be exerted to the priors. In addition, a lower cut off limit of $10^{-5}$ is necessary avoiding divided by zero. A visualization of priors can be found in Fig.~\ref{fig:6}.

\section{EXPERIMENTS}
In order to verify the model's performance and our proposed method to advance recall rate, we perform a series of experiments on two challenging benchmarks, \textbf{Cityscape} and \textbf{CamVid}, whose images have the resolutions of 2048 $\times$ 1024 and 960 $\times$ 720,  respectively.
\subsection{Implementation Details}
We use Tensorflow and a NVDIA GeForce GTX 1080Ti GPU for training and validation. We use Adam optimizer~\cite{kingma2014adam} to train our models with the initial learning rate set to 4$ \cdot $10$^{-4}$. We decay the learning rate with cosine annealing to the minimum  2.5$ \cdot $10$^{-3}$ of the initial learning rate in the final epoch. For the sake of combatting overfitting, we use the L2 weight regularization with decay of  1$ \cdot $10$^{-4}$. In view of the pre-training parameters by ImageNet\cite{russakovsky2015imagenet}, we update these parameters with 4 times smaller learning rate and weight decay. We scale the images with random factors between 0.75 and 1.25, and randomly crop the images. We use 768$\times$768 crops for Cityscapes, and 448$\times$448 crops for CamVid. Finally, we employ random horizontal flipping. We train our models with batch size 12, but when training models with IAL, the batch size decreases to 11 because of the limited resource. We train 200 epochs on Cityscapes, and 400 epochs on CamVid.
\begin{table}[t]
\caption{Results of different UDBs on Cityscapes (The mIoU is test at the resolution of 2048 $\times$ 1024. The FPS in calculated on NVDIA GeForce GTX 1080Ti at the resolution of 1024 $\times$ 512.)} 
\vskip-2ex
\label{tab:miou}
\begin{center}       
\begin{tabular}{|l|l|l|}
\hline
\rule[-1ex]{0pt}{3.5ex}  \textbf{UDB} & \textbf{mIoU} & \textbf{FPS}  \\

\hline
\rule[-1ex]{0pt}{3.5ex}  SwiftNet (Baseline)~\cite{orsic2019defense} & 69.8 & 78.4 \\

\hline
\rule[-1ex]{0pt}{3.5ex} ERF-PSPNet (Previous)~\cite{yang2018unifying}& 64.1 & 46.1  \\

\hline
\rule[-1ex]{0pt}{3.5ex}  ERF (Ours, dilation rate: 1,2,3) & 71.4 & 58.3  \\
\hline
\rule[-1ex]{0pt}{3.5ex}  ERF (Ours, dilation rate: 2,4,8) & 70.5 & 57.6  \\
\hline
\rule[-1ex]{0pt}{3.5ex} GCNET (Ours, late merging)& 70.8 & 61.2    \\
\hline 
\rule[-1ex]{0pt}{3.5ex} GCNET (Ours, early merging)& 72.2 & 58.6    \\
\hline

\end{tabular}
\end{center}
\vskip-2ex
\end{table}
\begin{table}[t]
\caption{CROSS-ENTROPY LOSS V.S IAL on CamVid (\%). (Pre stands for precision.)}
\vskip-2ex
\label{tab:camvid}
\begin{center}
\begin{tabular}{|c|c|c|c|c|c|c|c|c|c|c|c|}
\hline 
\multicolumn{1}{|c|}{\textbf{Group} } & \multicolumn{4}{c|}{\textbf{G3}} &\multicolumn{4}{c|}{\textbf{G2}} &\multicolumn{3}{c|}{\textbf{G1}}\\
\hline
\multicolumn{1}{|c|}{\textbf{Class}} & Sign & Car & Pedestrian & Bicyclist& Pole & Road & Sidewalk & Fence& Sky & Building & Tree\\
\hline
\multicolumn{12}{|c|}{\textbf{Cross-Entropy Loss}}\\
\hline
\rule[-1ex]{0pt}{3.5ex} \textbf{Pre} & 88.8& 95.8& 72.1& 85.5& 55.4& 97.7&91.1& 78.4& 96.7& 95.9& 84.0\\
\hline
\rule[-1ex]{0pt}{3.5ex} \textbf{Recall} & 45.9&95.7& 89.8& 71.5&66.8& 97.1& 94.2& 70.0& 95.9& 91.0& 91.7\\
\hline
\rule[-1ex]{0pt}{3.5ex} \textbf{IoU} & 43.4& 91.9& 66.7& 63.8& 43.4& 95.0& 86.2& 58.6& 92.9& 87.5& 78.0\\
\hline
\multicolumn{12}{|c|}{\textbf{IAL}}\\
\hline
\rule[-1ex]{0pt}{3.5ex} \textbf{Pre} & 79.9&95.5&70.3& 79.0&49.1& 97.5&\textbf{91.9}& 71.8& \textbf{97.4}& 95.4& 83.2\\
\hline
\rule[-1ex]{0pt}{3.5ex} \textbf{Recall} & \textbf{46.5}& \textbf{96.3}&  \textbf{90.1}& \textbf{77.9}& 63.3& \textbf{97.7}& 93.4& \textbf{71.2}& 94.5&90.0& \textbf{92.1}\\
\hline
\rule[-1ex]{0pt}{3.5ex} \textbf{IoU} & 41.6& \textbf{92.1}& 65.3& \textbf{64.6}& 38.2&  \textbf{95.3}& \textbf{86.3}& 55.6& 92.2& 86.3& 77.6\\
\hline
\end{tabular}
\end{center}
\vskip-2ex
\end{table}

\subsection{Quantitative Results}
In order to choose the best UDB structure as detailed in Sec.~\ref{sec:sfn}. We train four UDB architectures on Cityscapes training set, and evaluate them by using the Cityscapes validation set. In view of the fact that batch normalization layers can be fused with preceding convolution layers, we exclude them from the models when measuring the inference speed. The result and the inference speed on GTX 1080Ti are shown in Tab.~\ref{tab:miou} (The fps corresponds to the resolution of 1024 $\times$ 512). According to the result, we can draw the conclusion that our proposed factorized convolution UDBs can advance the basic performance, especially when using the early merging of GCNet, which obtains an mIoU improvement of roughly 2.4\%. In addition, we find that all our proposed models significantly surpass our previous ERF-PSPNet~\cite{yang2018unifying} in terms of both efficiency and accuracy. We can make an brief analysis among these UDBs. The ERF version of UDB is effective, but the performance depends on the chosen dilation rates. We find that using larger dilation rates results in worse performance than smaller one. This is because unlike ERFNet which is composed of a series of factorized convolution blocks, our model only possesses three UDBs so that using excessively big dilation rates produces sparse connection on the feature maps. Therefore, using smaller dilation rates can enlarge the valid receptive field and advance the performance. As for the GCNet block, we find that the early merging version is much better than the late merging. The feature from encoder is low-level features, and the late merging of GCNet operates on the features may lack the relation between the low-level features and high-level features from the previous layers. However, the early merging of GCNet utilizes element-wise summations to merge these features, followed by the GCNet to extract the features and enlarge valid receptive field. Therefore, we select the early merging of GCNet as the baseline for the following experiments. Additionally, the result on CamVid tesing set shows that our proposed method can prove its effectiveness, \textit{i.e.} 73.4\% over 72.3\% for the basic version according to Tab.~\ref{tab:camvid}.

\begin{table}[t]
\caption{CROSS-ENTROPY LOSS V.S IAL on Cityscapes for G3 (\%). }
\vskip-2ex
\label{tab:city1}
\begin{center}
\begin{tabular}{|c|c|c|c|c|c|c|c|c|}
\hline 
\textbf{Class} & Sign & Rider & Truck & Bus& Train & Motorcycle& Bicycle& Traffic light\\
\hline
\multicolumn{9}{|c|}{\textbf{Cross-Entropy Loss}}\\
\hline
\rule[-1ex]{0pt}{3.5ex} \textbf{Pre} &77.7& 73.4& 86.1& 89.9& 86.7& 71.6&79.2& 69.3\\
\hline
\rule[-1ex]{0pt}{3.5ex} \textbf{Recall} & 90.4&72.5& 79.1&88.6&76.4&64.9& 90.9& 88.9\\
\hline
\rule[-1ex]{0pt}{3.5ex} \textbf{IoU} & 71.8& 57.4& 70.1&80.5& 68.3& 51.6& 73.4& 63.8\\
\hline
\multicolumn{9}{|c|}{\textbf{IAL}}\\
\hline
\rule[-1ex]{0pt}{3.5ex} \textbf{Pre} & 61.5&66.3&75.8& 80.3&72.0&68.2&72.1&53.9\\
\hline
\rule[-1ex]{0pt}{3.5ex} \textbf{Recall} & \textbf{93.5}& \textbf{76.1}&  73.3& \textbf{90.6}& \textbf{82.8}& \textbf{68.0}& \textbf{92.9}& \textbf{92.3}\\
\hline
\rule[-1ex]{0pt}{3.5ex} \textbf{IoU} & 59.0& 54.9&59.4& 74.1& 62.6&  51.6& 68.3& 51.6\\
\hline
\end{tabular}
\end{center}
\vskip-2ex
\end{table}
\begin{table}[t]
\caption{CROSS-ENTROPY LOSS V.S IAL on Cityscapes for G2 and G1 (\%).}
\vskip-2ex
\label{tab:city2}
\begin{center}
\begin{tabular}{|c|c|c|c|c|c|c|c|c|c|c|c|}
\hline 
\multicolumn{1}{|c|}{\textbf{Group} } & \multicolumn{5}{c|}{\textbf{G2}} &\multicolumn{6}{c|}{\textbf{G1}} \\
\hline
\multicolumn{1}{|c|}{\textbf{Class}} & Car & Sidewalk & Fence & Pole& Pedestrian & Road & Building & Wall& Tree & Terrain & Sky\\
\hline
\multicolumn{12}{|c|}{\textbf{Cross-Entropy Loss}}\\
\hline
\rule[-1ex]{0pt}{3.5ex} \textbf{Pre} & 95.9& 86.1&76.4& 65.3& 78.8& 99.0&96.1& 73.9& 95.4& 76.0&93.7\\
\hline
\rule[-1ex]{0pt}{3.5ex} \textbf{Recall} & 97.6&90.6& 65.8& 83.5&94.5& 97.9& 93.4& 51.3&95.2& 73.6& 98.6\\
\hline
\rule[-1ex]{0pt}{3.5ex} \textbf{IoU} & 93.6& 79.0& 54.7& 57.8&75.3& 96.9& 90.0& 43.4& 91.0& 59.7& 92.4\\
\hline
\multicolumn{12}{|c|}{\textbf{IAL}}\\
\hline
\rule[-1ex]{0pt}{3.5ex} \textbf{Pre} & 94.0&82.5&65.8& 53.4&76.0& \textbf{99.1}&\textbf{96.7}& 68.4& \textbf{95.7}& 66.1& 92.7\\
\hline
\rule[-1ex]{0pt}{3.5ex} \textbf{Recall} & \textbf{97.7}& \textbf{91.3}&  \textbf{69.7}& \textbf{84.9}& 94.0& 97.0& 89.6& 46.6& 92.8&77.5& 97.7\\
\hline
\rule[-1ex]{0pt}{3.5ex} \textbf{IoU} & 92.0& 76.4& 51.2& 48.8& 72.4&  96.1& 86.9&38.3& 89.1& 55.5& 90.7\\
\hline
\end{tabular}
\end{center}
\vskip-2ex
\end{table}
\begin{table}[t]
\caption{Result of GCN on Cityscapes (\%).}
\vskip-2ex
\label{tab:city3}
\begin{center}
\begin{tabular}{|c|c|c|c|c|c|c|c|c|c|c|}
\hline 
\rule[-1ex]{0pt}{3.5ex}\textbf{Class} & Sign & Rider & Truck & Bus& Train & Terrain& Bicycle& Traffic light&Car&Sidewalk\\
\hline
\rule[-1ex]{0pt}{3.5ex} \textbf{Pre} & \textbf{77.9}& 72.8&86.5& 85.5& 81.4& 72.5&78.4&  \textbf{70.0}& 95.8&86.0\\
\hline
\rule[-1ex]{0pt}{3.5ex} \textbf{Recall} & 90.1&\textbf{72.7}& \textbf{80.7}& \textbf{90.1}& 68.9&  \textbf{74.3}& \textbf{91.3}& \textbf{90.1}&\textbf{97.6}&90.5\\
\hline
\rule[-1ex]{0pt}{3.5ex} \textbf{IoU} & 71.7& 57.2& \textbf{71.7}& 78.1&59.5& 58.0& 73.0&  \textbf{64.9}&\textbf{93.6}& 78.8\\
\hline
\rule[-1ex]{0pt}{3.5ex}\textbf{Class} & Fence & Pole& Wall & Road & Building & Pedestrian& Tree & Motorcycle & Sky&\textbf{Mean}\\
\hline
\rule[-1ex]{0pt}{3.5ex} \textbf{Pre} &71.4&\textbf{65.9}&\textbf{77.2}& 99.0&\textbf{96.1}& \textbf{80.0}&95.3& 71.5&\textbf{94.2}&82.0\\
\hline
\rule[-1ex]{0pt}{3.5ex} \textbf{Recall} & \textbf{68.7}& 83.5&  47.5& 94.4& 93.3&97.8&95.1&61.8& 98.0&83.5\\
\hline
\rule[-1ex]{0pt}{3.5ex} \textbf{IoU} & 53.8& \textbf{58.3}& 41.7& 96.8&89.9& \textbf{76.4}&90.9&49.6&92.4& 71.4\\
\hline
\end{tabular}
\end{center}
\vskip-2ex
\end{table}
\begin{table}[t]
\caption{ML rule on Cityscapes (\%).}
\vskip-2ex
\label{tab:city4}
\begin{center}
\begin{tabular}{|c|c|c|c|c|c|c|c|c|c|c|}
\hline 
\rule[-1ex]{0pt}{3.5ex}\textbf{Class} & Sign & Rider & Truck & Bus& Train & Terrain& Bicycle& Traffic light&Car&Sidewalk\\
\hline
\rule[-1ex]{0pt}{3.5ex} \textbf{Pre} & 47.3& 43.5&57.8& 69.5& 36.7& 39.8&63.7& 30.7& 94.4& 82.4\\
\hline
\rule[-1ex]{0pt}{3.5ex} \textbf{Recall} & \textbf{94.2}&\textbf{80.1}& \textbf{83.0}& \textbf{91.9}& \textbf{82.5}&  \textbf{77.1}& 90.0& \textbf{95.4}&95.3& 88.5\\
\hline
\rule[-1ex]{0pt}{3.5ex} \textbf{IoU} & 46.0& 39.2& 51.6& 65.5&34.0& 35.6& 59.5& 30.3& 90.1& 74.4\\
\hline
\rule[-1ex]{0pt}{3.5ex}\textbf{Class} & Fence & Pole& Wall & Road & Building & Pedestrian& Tree & Motorcycle & Sky&\textbf{Mean}\\
\hline
\rule[-1ex]{0pt}{3.5ex} \textbf{Pre} & 57.5&37.6&55.4& 98.6&\textbf{96.4}& 65.0&95.4& 31.6& 87.2& 62.6\\
\hline
\rule[-1ex]{0pt}{3.5ex} \textbf{Recall} & \textbf{69.2}& \textbf{88.8}&  \textbf{63.8}& 93.5& 82.3& 93.5& 88.0&\textbf{76.7}& 94.6&\textbf{85.7}\\
\hline
\rule[-1ex]{0pt}{3.5ex} \textbf{IoU} & 45.8& 35.9& 42.1& 92.2& 79.8&  62.2&84.5&28.9& 83.0& 56.9\\
\hline
\end{tabular}
\end{center}
\vskip-2ex
\end{table}

In the following experiments, our proposals including importance-aware loss function, GCN and ML rule are analyzed quantitatively.

\textbf{Importance-aware loss:} As detailed in Sec.~\ref{sec:intro}, in order to promote the recall rate for important classes, we need to category classes into different importance groups. For CamVid, we select the ranks of importance as depicted in Fig.~\ref{fig:4}(a). From the results in Tab.~\ref{tab:camvid}, we observe that the recall rates of G3 have been advanced especially the bicyclist which has an improvement of around 6.4\%. And the mean recall rate rises to 83.0\% from 82.7\%, while the G3's mean recall rate rises to 81.7\% from 80.0\%. In other words, IAL make the model focus on the detection of the most important categories successfully, evidenced by the improvement of the important categories' recall rates. In view of the difference between Cityscapes and CamVid, some categories need to be attributed with different importance ranks or else it may lead to detrimental effects for training. For Cityscapes, we regard the importance categories as shown in Tab.~\ref{tab:city1} and Tab.~\ref{tab:city2}. We find that the G3's recall rates have been improved dramatically, especially the train, which has a improvement of 6.4\%. We observe that the mean recall rate, G3's mean recall rate and G2's mean recall rate have elevated to 84.6\% from 83.9\%,  83.7\% from 81.5\% and 87.5\% from 86.4\%, respectively. In addition, many G1's categories have a slight improvement of precision. Making a brief summary to IAL, it can indeed advance the model's recall rate, especially the important categories' recall rates, but may have an adverse effect on the mIoU depending on the dataset. For CamVid, the mIoU has a slight degradation from 73.4\% to 72.3\%, but mIoU of Cityscapes dramatically degrades from 72.2\% to 67.3\%.

\textbf{GCN:} GCN's performance depends on the pre-defined graph. In our experiment, we train models on Cityscapes, and define the graph like importance group. We regard that the G3's nodes have connection with every nodes, G2's nodes have connection with every nodes except G3's nodes, and G1's nodes only have dependency with G1's nodes. We select one-hot encoding label as the word embeddings for the nodes. The result of GCN is filled in the Tab.~\ref{tab:city3}. In comparison with the results on Tab.~\ref{tab:city1} and Tab.~\ref{tab:city2}, we observe that GCN advance many important categories of G2 and G3 's recall rate, but the advancement is not remarkable, and it can improve some categories' precision as well. In general, it can indeed improve the important categories' recall rates. In addition, it does not have a severe side effect on the precision or the IoU of the model. On the other hand, a more scientific method to produce graph is promising to advance the model's performance, and other types of word embeddings may improve it as well.

\textbf{Bayes \textit{V.S.} Maximum Likelihood (ML) decisions:} As detailed in Sec.~\ref{sec.dr}, we obtain the priors by calculating the pixel-wise class distribution of the Cityscapes training set. A lower cut off limit of $10^{-5}$ and a Gaussian filter with $\sigma$ of 40 to avoid being divided by zero and reduce noise. The results in Tab.~\ref{tab:city4} shows that ML rule has an improvement on recall rate by a large margin. The recall rates of some categories like wall, motorcycle and rider have a remarkable elevation. On the contrary, the precision of the model has a detrimental effect, and the mean precision degrade from 82.7\% to 62.6\%, in spite of the fact that the mean recall rate has an improvement of around 1.8\%. Comparing with Tab.~\ref{tab:city1} and Tab.~\ref{tab:city2}, we draw the conclusion that the categories with improvement on recall rate are the categories with relative lower precision of the model trained by using the cross-entropy loss.

Making a brief comparison among these three methods for high recall rate, we conclude that our three proposed methods can advance the recall rates, but all of them have a side effect on precision. For IAL, it can push the model to focus on the important categories, and detect them with high recall rates. It is promising to be taken into training the model for autonomous vehicles because categories which need to be detected with high recall rates like bus, car, pedestrian, and etc,. are the most important key to offer a safe path for the vehicles even if their precisions are comparatively low. Moreover, IAL can be modified into other versions combined with focal loss and other methods. For GCN, we find it has the least side effect on the performance of model. Due to the fact that GCN is utilized into extract correlation among nodes or classes, it could theoretically have a better performance than traditional classifier based on CNNs. However, the pre-defined graph influences the result. A more scientific method for searching the graph is the key point. A possible solution to this issue is to construct a learnable graph module to search for the best graph for GCN other than define the graph by ourselves. For another thing, a better word embedding needs to be considered. For ML rule, an amazing improvement on recall rate can not fade away the detrimental decrease of precision. There are many possible keys to cope with the problem. Our training for ML rule is the same as Bayes rule. In other words, we merely train the model depending on max Bayes probability other than conditional probability. Training with priors may be a workable way.

\subsection{Qualitative Analysis}
\begin{figure}[t] 
  \centering 
  \subfigure[Image]{ 
    \includegraphics[height=2.9cm]{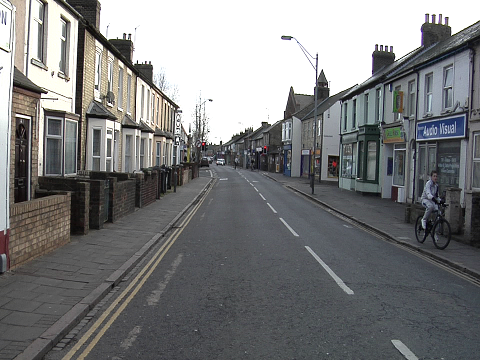} 
  } 
  \subfigure[Label]{ 
    \includegraphics[height=2.9cm]{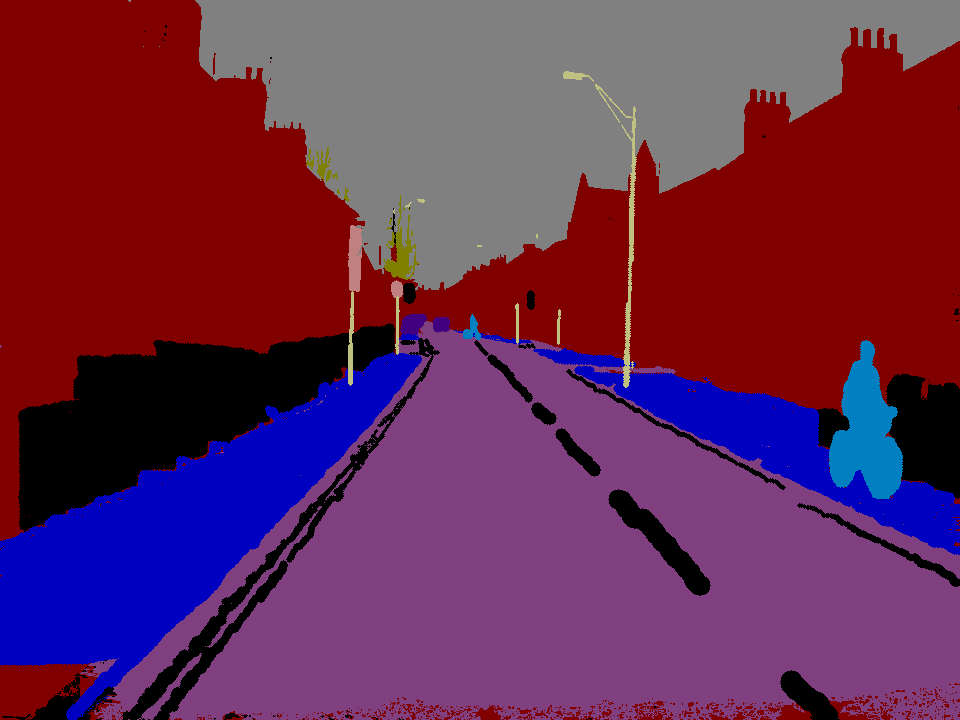} 
  } 
    \subfigure[Baseline]{ 
    \includegraphics[height=2.9cm]{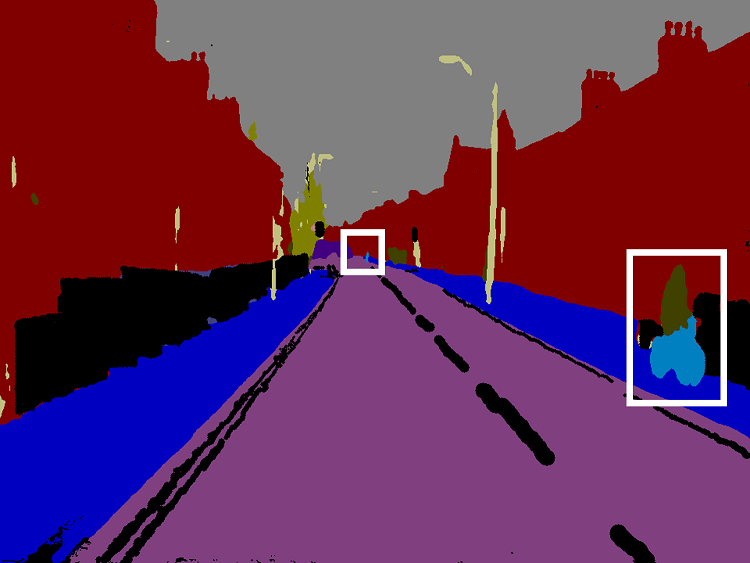} 
  } 
    \subfigure[IAL]{ 
    \includegraphics[height=2.9cm]{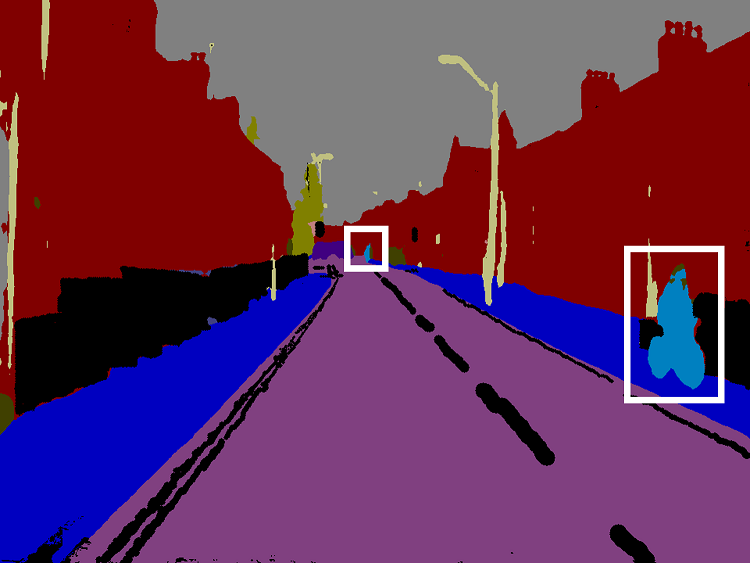} 
  } 
  \caption{\label{fig:camial}The result comparison between baseline and IAL on CamVid. The white box area is the main difference area, and the black area is the mask where pixels can be ignored.} 
\end{figure}   

\begin{figure}[t] 
  \centering 
  \subfigure[Image]{ 
    \includegraphics[height=2.9cm]{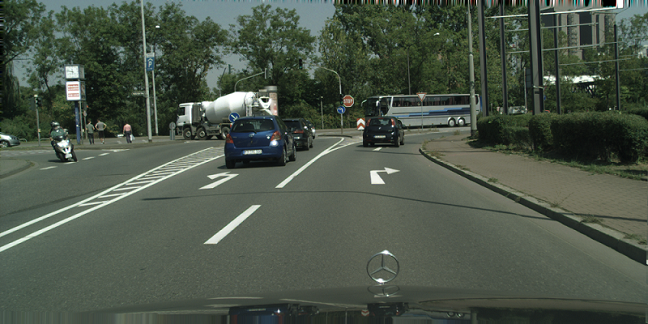} 
  } 
  \subfigure[Label]{ 
    \includegraphics[height=2.9cm]{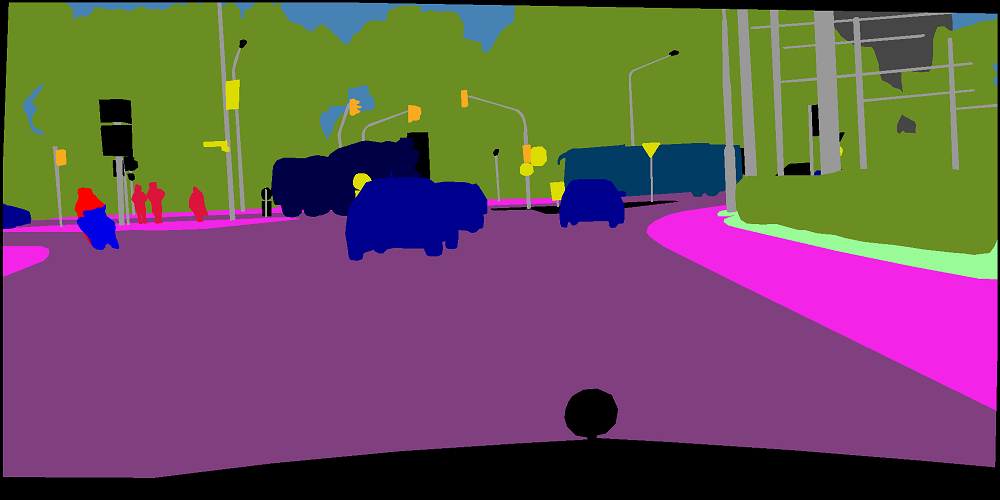} 
  } 
    \subfigure[Baseline]{ 
    \includegraphics[height=2.9cm]{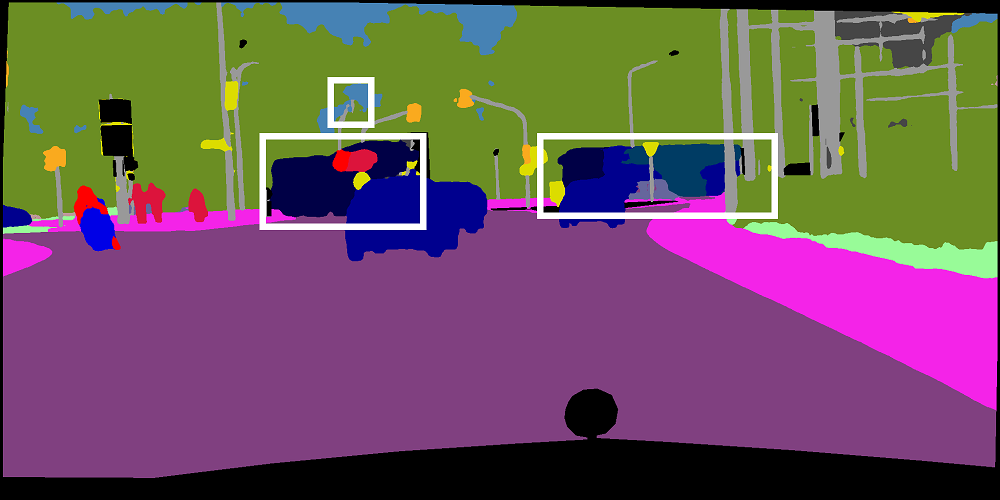} 
  } 
    \subfigure[IAL]{ 
    \includegraphics[height=2.9cm]{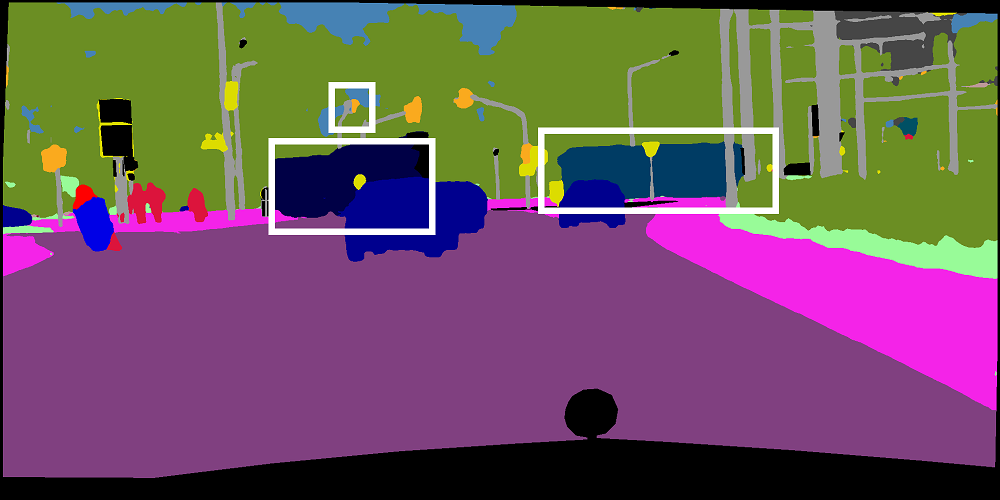} 
  } 
  \caption{\label{fig:cityial}The result comparison between baseline and IAL on Cityscapes.} 
\end{figure}

\begin{figure}[t] 
  \centering 
  \subfigure[Image]{ 
    \includegraphics[height=2.9cm]{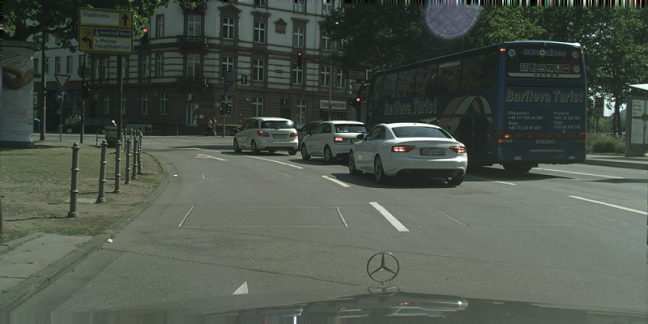} 
  } 
  \subfigure[Label]{ 
    \includegraphics[height=2.9cm]{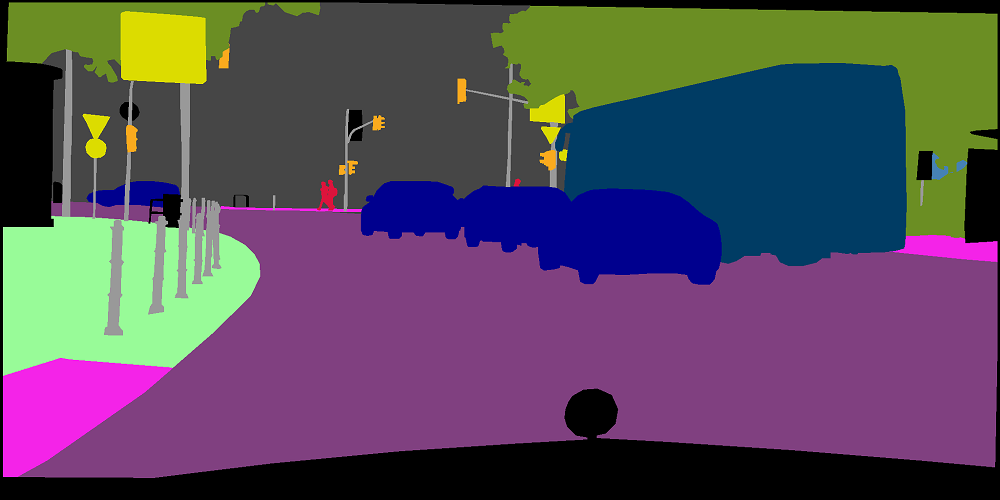} 
  } 
    \subfigure[Baseline]{ 
    \includegraphics[height=2.9cm]{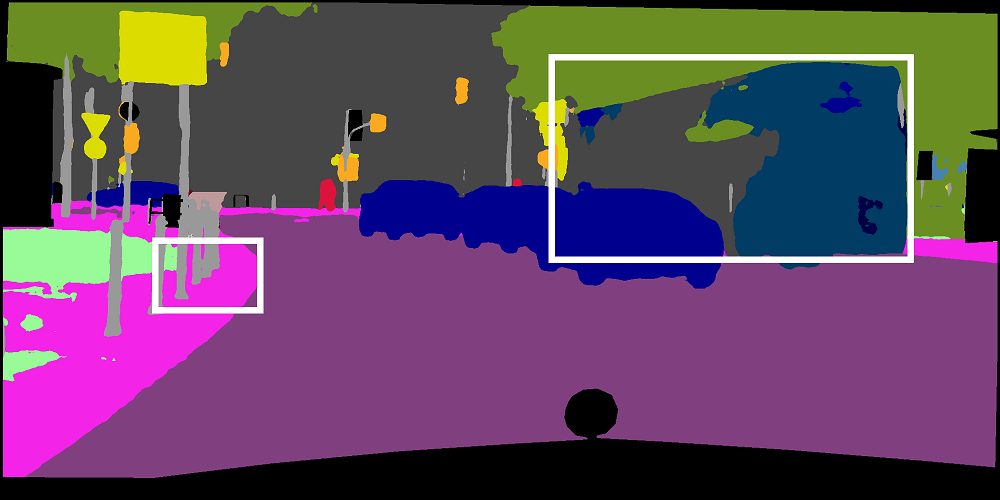} 
  } 
    \subfigure[GCN]{ 
    \includegraphics[height=2.9cm]{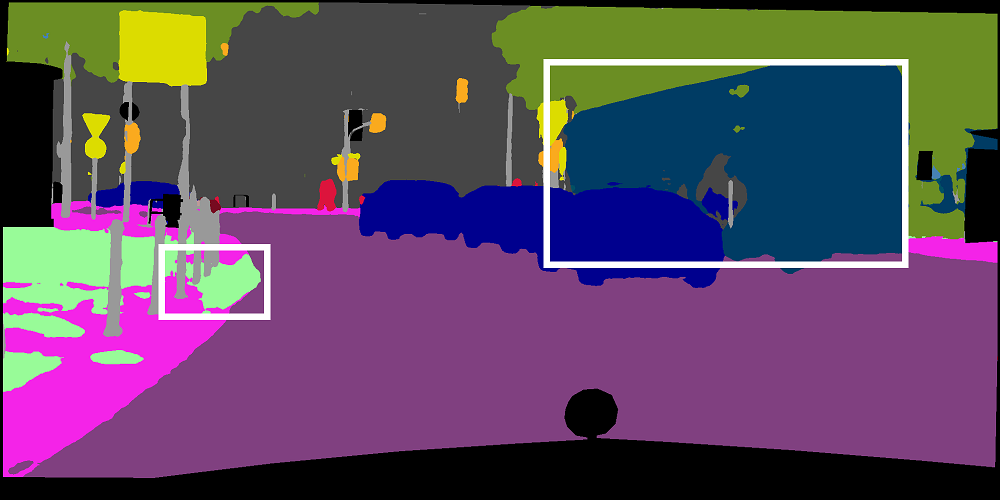} 
  } 
  \caption{\label{fig:citygnn}The result comparison between baseline and GCN on Cityscapes.} 
\end{figure} 

   \begin{figure} [t]
   \begin{center}
   \includegraphics[height=5cm,width=10cm]{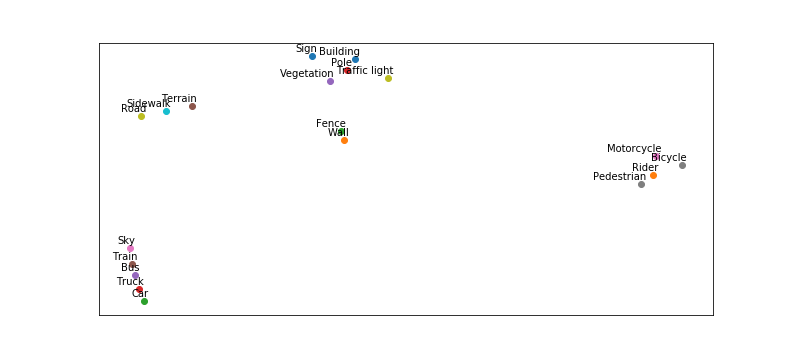}
   \end{center}
   \vskip-4ex
   \caption[example] 
   { \label{fig:tsne}t-SNE on the GCN classifier. The closer the items are, the closer the semantic meaning is. It shows the inter-dependent relationship among these items: train, bus, truck and car are close to each other, while motorcycle, bicycle, rider and pedestrian are close to each other.}
   \end{figure} 

\begin{figure}[t] 
  \centering 
  \subfigure[Image]{ 
    \includegraphics[height=2.9cm]{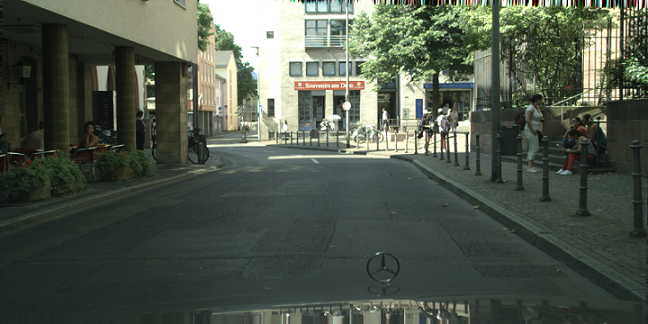} 
  } 
  \subfigure[Label]{ 
    \includegraphics[height=2.9cm]{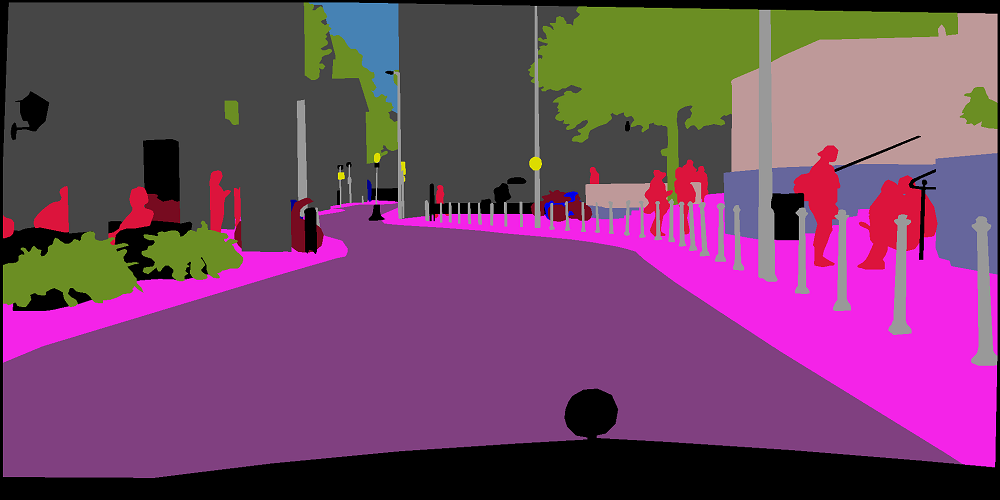} 
  } 
    \subfigure[Baseline]{ 
    \includegraphics[height=2.9cm]{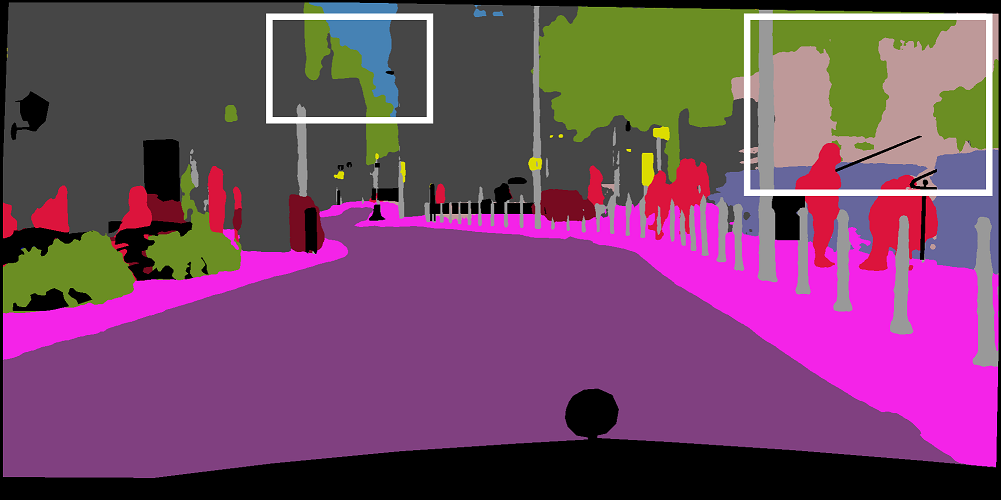} 
  } 
    \subfigure[ML]{ 
    \includegraphics[height=2.9cm]{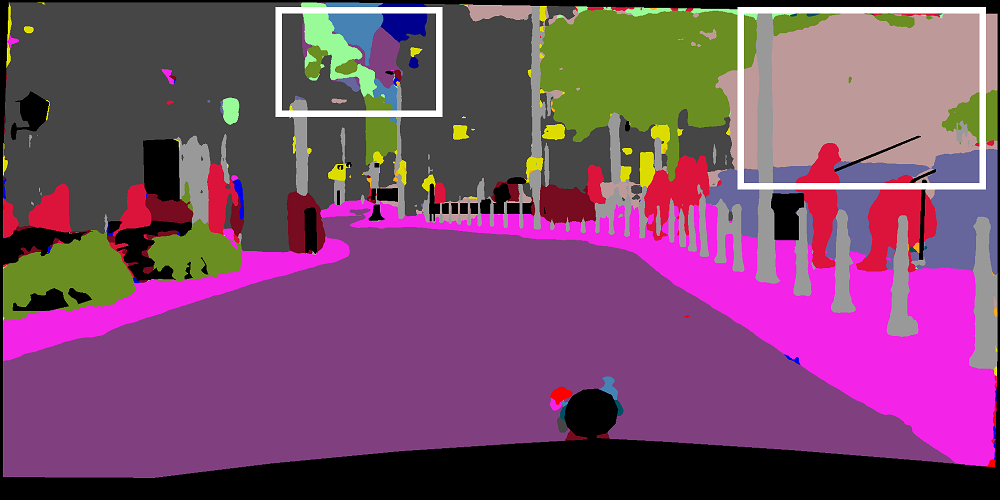} 
  } 
  \caption{\label{fig:cityml}The result comparison between baseline and ML decision rule on Cityscapes.} 
\end{figure} 
The effects of IAL are illustrated in Fig.~\ref{fig:camial} and Fig.~\ref{fig:cityial}. For the former, we find that the bicyclist can be detected with high recall rate, and the model trained with cross-entropy loss misses many parts of the bicyclist and even ignore the bicyclist in the distance. For the latter, the model trained with cross-entropy loss segments part of the bus into bus, car, and truck, what's more, it even categories the truck into pedestrian as labeled by red, and it ignores traffic light which is successfully detected by using IAL. However, the model trained by cross-entropy loss possesses a higher precision on the area like vegetation and terrain.

In order to prove the effectiveness of GCN, we utilize t-SNE~\cite{maaten2008visualizing} to visualize the effect of the GCN classifier. The learned parameters are shown in Fig.~\ref{fig:tsne}, we observe that the classifier maintains meaningful semantic topology, which puts car, truck, bus and train close and motorcycle, bicycle, rider and pedestrian together and etc. Therefore, the classifier can select meaningful features for classification. The result of the SS is shown in Fig.~\ref{fig:citygnn}. We find that the baseline segments a large part of the bus into pole, part of the terrain into sidewalk. On the contrary, the model with GCN classifier detects most part of the bus and has a higher recall rate than the baseline method on the area of terrain. We believe a more scientific graph is promising for better performance.

For ML decision rule, it is obvious to find that the result of using ML decision rule segments most part of wall, while the result of Bayes decision rule, segments many part of wall into vegetation as shown in Fig.~\ref{fig:cityml}. However, the result conveys the fact that ML decision rule can bring in much noise, especially in the white box of the Fig.~\ref{fig:cityml} where the result of using Bayes decision rule segments vegetation resoundingly while the ML rarely ignores any part of it. On the other hand, we find that the boundary of the object is partial to be segmented into objects that are relatively rare. In other words, the ML decision rule is more likely to segment boundary pixels to the objects that are detected with low precision when using Bayes rule.

\section{CONCLUSIONS AND FUTURE WORK}
Semantic Segmentation (SS) can bring brilliant fruits in many aspects of computer vision in the era of AI, especially autonomous driving and medical image processing. For certain application, certain evaluation criterion should be taken into account. In our work, our main purpose is to advance the recall rate of the model especially certain categories, because in the area of autonomous driving, different object possess different ranks of importance, and many important items should be detected and segmented with high recall rates. We put forward there different methods to attain higher recall rates. In spite of the fact these three methods can indeed advance recall rates to some extent, there are still much space for a better performance following these three methods.

In order to put SS into practice, we must take inference speed into account. We adapt SwiftNet into Swift Factorized Network (SFN) combining GCNet and factorized convolution which possesses a larger valid receptive field. In the future, we aim to further explore the proposed methods towards higher recall rate, especially GCN. We have the intention to design a better scheme to search for a better pre-defined graph and word-embeddings instead of using one-hot encoding merely. Moreover, training a model by using conditional probability instead of Bayes probability is a possible solution. Furthermore, we are determined to combine SS~\cite{yang2018unifyingterrainawareness}\cite{yang2019can} and visual localization~\cite{lin2018visual}\cite{cheng2019panoramic} to serve autonomous vehicles and navigation assistance systems for the visually impaired.
\bibliography{report} 
\bibliographystyle{spiebib} 

\end{document}